\pdfoutput=1
\documentclass[letterpaper]{article} 
\usepackage{aaai2026}  
\usepackage{times}  
\usepackage{helvet}  
\usepackage{courier}  
\usepackage[hyphens]{url}  
\usepackage{graphicx} 
\usepackage{natbib}  
\usepackage{caption} 
\usepackage{multirow}
\usepackage{booktabs} 
\usepackage{makecell} 
\usepackage[switch]{lineno}

\usepackage{marvosym} 

\usepackage{algorithm}
\usepackage{algorithmic}
\usepackage{amsmath}
\usepackage{newfloat}
\usepackage{listings}
\DeclareCaptionStyle{ruled}{labelfont=normalfont,labelsep=colon,strut=off} 
\floatstyle{ruled}
\newfloat{listing}{tb}{lst}{}
\floatname{listing}{Listing}
\title{Interpretable Oracle Bone Script Decipherment through Radical and Pictographic Analysis with LVLMs}

\author{
    Kaixin Peng\textsuperscript{\rm 1}\textsuperscript{*},
    Mengyang Zhao\textsuperscript{\rm 1}\textsuperscript{*},
    Haiyang Yu\textsuperscript{\rm 1}\textsuperscript{*}\textsuperscript{\dag},
    Teng Fu\textsuperscript{\rm 1, 2},
    Bin Li\textsuperscript{\rm 1}\textsuperscript{\Letter}
}
\affiliations{
    \textsuperscript{\rm 1}Fudan University \quad
    \textsuperscript{\rm 2}ByteDance Inc.\\
}

\makeatletter
\def\@copyrightspace{}
\makeatother

\usepackage{bibentry}
\usepackage{xcolor}

\begin{document}
\nocopyright
\maketitle

\begingroup
\renewcommand\thefootnote{}%
\footnotetext{* Equal contribution, \dag\ Project lead, \Letter\ Corresponding author.}%
\endgroup

\begin{abstract}

As the oldest mature writing system, Oracle Bone Script (OBS) has long posed significant challenges for archaeological decipherment due to its rarity, abstractness, and pictographic diversity. Current deep learning-based methods have made exciting progress on the OBS decipherment task, but existing approaches often ignore the intricate connections between glyphs and the semantics of OBS. This results in limited generalization and interpretability, especially when addressing zero-shot settings and undeciphered OBS. To this end, we propose an interpretable OBS decipherment method based on Large Vision-Language Models, which synergistically combines radical analysis and pictograph-semantic understanding to bridge the gap between glyphs and meanings of OBS. Specifically, we propose a progressive training strategy that guides the model from radical recognition and analysis to pictographic analysis and mutual analysis, thus enabling reasoning from glyph to meaning. We also design a Radical-Pictographic Dual Matching mechanism informed by the analysis results, significantly enhancing the model's zero-shot decipherment performance. To facilitate model training, we propose the Pictographic Decipherment OBS Dataset, which comprises 47,157 Chinese characters annotated with OBS images and pictographic analysis texts. Experimental results on public benchmarks demonstrate that our approach achieves state-of-the-art Top-10 accuracy and superior zero-shot decipherment capabilities. More importantly, our model delivers logical analysis processes, possibly providing archaeologically valuable reference results for undeciphered OBS, and thus has potential applications in digital humanities and historical research. The dataset and code will be released in \textcolor{blue}{\url{https://github.com/PKXX1943/PD-OBS}}.

\end{abstract}

\section{Introduction} 
\begin{figure*}[t]
  \centering
  \includegraphics[width=0.96 \textwidth]{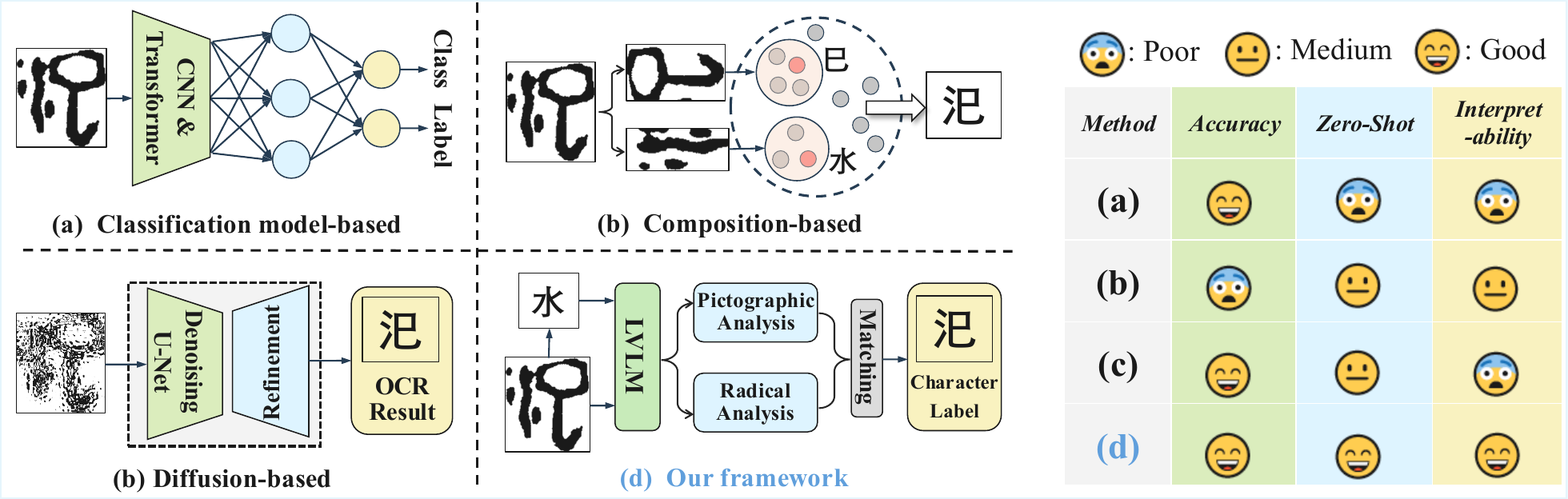}
  \caption{Illustrations of the three existing paradigms and our proposed framework.
Our method is able to take account of classification accuracy, zero-shot decipherment, and interpretability and achieving better performance in all of them.}
  \label{fig:intro}
\end{figure*}

Oracle Bone Script (OBS) represents the earliest known mature writing system, primarily used between the 14th and 11th centuries BCE, making it invaluable for archaeology and history. These characters, inscribed on turtle plastrons and animal bones, feature concise, fluid strokes and highly pictographic forms that often resemble the shapes of real-world objects. Traditional decipherment relies on expert knowledge and manual efforts, making it slow and hard to scale for the growing corpus of script. Recently, deep learning methods have drawn attention and shown promising progress in OBS decipherment.

Nevertheless, deciphering OBS remains a formidable challenge due to the rarity, abstraction, and diversity of its glyphs, as well as the lack of complete contextual information. To date, over 4,500 unique oracle bone characters have been discovered, yet less than one-third have been successfully deciphered. 
Early classification model-based approaches \cite{inception-v3,MAAN_cnn-based,swin_cnn-based,PYGT,orerf-cnn_based,oraclepoints} primarily relied on CNN or Transformer-based visual backbones to perform closed-set classification of oracle bone characters.
Despite their effectiveness, these methods are confined to the closed-set setting, making them inapplicable to zero-shot setting and undeciphered characters and thus significantly limiting their scope of application. 
In recent years, composition-based methods \cite{cola_composition-based,ppp_composition-based,retrieval_composition-based,seg_composition-based} have been proposed to solve deciphering within open-set. These method exhibits a certain degree of zero-shot capability and interpretability, however, it overlooks the rich associations between pictographic forms and semantic information inherent to OBS, which leads to relatively low decipherment accuracy.
In addition, diffusion-based methods \cite{obsd, bbdm} for OBS decipherment have been proposed, achieving significant advances in both accuracy and zero-shot capability through conditional control and sampling strategies. Unfortunately, the instability of these methods results in low interpretability and reliability of the decipherment outputs, thereby limiting their practicality in real-world applications.

Multiple studies on OBS \cite{pic_1, pic_2} and Chinese \cite{yu_radical,yu_clip} have demonstrated that the semantic information conveyed by radical glyphs often determines the fundamental meaning of a character, and that the pictographs are also highly correlated with semantic contents. Therefore, we consider that leveraging both radical and pictographic information may significantly enhance the model’s ability to recognize OBS and interpret the recognition process, which is especially valuable for undeciphered OBS.
To this end, we propose to bridge the glyphs and meanings of OBS using the powerful cross-modal reasoning ability of Large Vision-Language Models (LVLMs). Specifically, we leverage LVLMs to analyze the semantic meaning of both the radical and the overall character based on the pictographic features. By combining these two aspects, we motivate the model to obtain comprehensive OBS meanings and find suitable decipherment candidate sets based on the meanings. This allows our model to cope with undeciphered OBS and explain the logical analysis chains from glyphs to the meaning of OBS, enhancing the interpretability and generalization of the decipherment process. 


Although LVLMs have achieved excellent performance on many tasks \cite{GOT,chatreid}, it is still difficult to apply directly to decipherment tasks due to lack of the domain-specific knowledge of OBS. To address this, we introduce a pictographic analysis dataset PD-OBS and a progressive training strategy to specialize LVLMs for OBS decipherment. The PD-OBS dataset contains approximately 50,000 Chinese characters based on the Kangxi Dictionary and deciphered OBS. Each character is associated with oracle bone images, modern and ancient scripts, and is annotated with detailed radical analysis and pictographic meaning labels, providing comprehensive support for the decipherment framework. Regarding the progressive training strategy, we first train the model to perform radical recognition and analyze the semantic information embedded to get the meaning of the underlying characters determined by the radicals. Then, we train the model to perform pictographic analysis on the whole character to grasp the character-level semantic meanings. Finally, we utilize the mutual analysis so that the two levels of analysis complement each other. In addition, we propose a novel Radical-pictographic Dual Matching mechanism, which uses the analysis results to find suitable candidate characters in the dictionary and brings better zero-shot performance. Experiments demonstrate that our method achieves more accurate decipherment with excellent zero-shot capability and decipherment interpretation. The main contributions of this work are as follows:
\begin{itemize}
    \item We propose an LVLM-based decipherment framework to bridge the pictographic and semantic of OBS, which is the first method to try explaining the decipherment process and handling undeciphered scripts.
   
   \item  We designed a progressive training to gradually guide the model in building relationships between glyphs and meanings through radical recognition and analysis, pictogram analysis, and mutual analysis. Based on the intermediate results, we designed a novel Radical-pictographic Dual Matching mechanism to replace directly deciphering, thus achieving better performance, especially at zero shots. 
    \item We propose the PD-OBS dataset, which is the first large-scale resource including both character structure analysis and pictographic analysis annotation.
    \item  Our method achieves state-of-the-art Top-10 accuracy and strong zero-shot ability on public decipherment benchmarks and can offer plausible reference results for previously undeciphered oracle characters.
\end{itemize}

\section{Related Work}
\subsection{Oracle Bone Script Datasets}
With the continuous excavation of OBS and the steady expansion of digitized resources, an increasing number of high-quality datasets \cite{hwobc,seg_composition-based,oracle-50k,obc360,obi125,survey,survey2,survey3} have been curated and released as open-access resources. Since the introduction of the first publicly available OBS dataset Oracle-20K \cite{oracle-20k}, the volume and quality of data have improved significantly.
In particular, the release of two large datasets, HUST-OBC \cite{hust-obs} and EVOBC  \cite{evobc}, has greatly expanded the pool of available data. These datasets contain a total of more than 70,000 oracle bone character samples covering more than 3,000 different Chinese character categories. 

Currently, HUST-OBC~\cite{hust-obs} and EV-OBC \cite{evobc} are the most widely adopted benchmark datasets for OBS research. The HUST-OBC dataset, derived from books, websites, and the collation of previous datasets, collects 77064 sample scanned or handwritten images of a total of 1588 deciphered character classes, as well as 62989 scanned images of undeciphered samples. The EV-OBC dataset contains 229,170 images collected from authoritative literature and websites, containing a total of 13,714 different character classes. These images cover six historical stages of ancient scripts, namely: OBS, Chinese Bronze Inscriptions, Seal Script, 
Spring and Autumn Period Script, Warring States Period script, and Clerical Script, where OBS account for about one-third.

\subsection{Oracle Bone Script Decipherment}

Recently, classification model-based approaches \cite{FLR_pr,TLC_pr,swin_cnn-based,orerf-cnn_based,oraclepoints, 2dl_cnn,ViT,survey, survey2, survey3} for OBS decipherment have emerged, achieving performance on closed-set tasks that matches or even surpasses that of human archaeological experts. For example, Enhanced Inception-V3\cite{inception-v3} employs convolutional attention modules instead of standard convolutional layers to improve decipherment performance based on a CNN backbone. Building on Transformer architectures, the Pyramid Graph Transformer\cite{PYGT} integrates a pyramid-structured Vision Transformer (ViT) with skeleton graph representations, attaining state-of-the-art results in closed-set OBS decipherment.

In the open-set scenario, Wang \textit{et al.}~\cite{ppp_composition-based} attempted to decompose the structural components of OBS using segmentation models, followed by clustering methods to align these components with the radicals of modern Chinese characters. Although this method facilitates interpretability and archaeological validation, it fails to account for the significant differences in glyph structure between OBS and modern Chinese, resulting in limited decipherment accuracy. In addition, the OBSD\cite{obsd} method, which is based on diffusion models, combines local structure sampling with style adaptation to establish effective correspondences between OBS and modern Chinese characters, using OBS as a conditional input to guide Chinese character generation and achieving impressive accuracy. However, this method suffers from instability, unpredictable output, and a lack of interpretability. Oraclesage \cite{oraclesage} is the first to employ LVLM for the description and analysis of OBS, but its accuracy remains suboptimal, primarily due to the insufficient exploitation of glyph features and existing dictionary resources. Therefore, we propose to bridge the glyphs and meaning based on LVLMs and propose a large pictographic decipherment dataset to adapt LVLMs to OBS, with the aim of enhancing the accuracy, generalization, and interpretability of OBS decipherment.

\section{Pictographic Decipherment OBS Dataset}

\begin{figure}[]
  \centering
  \includegraphics[width=0.49 \textwidth]{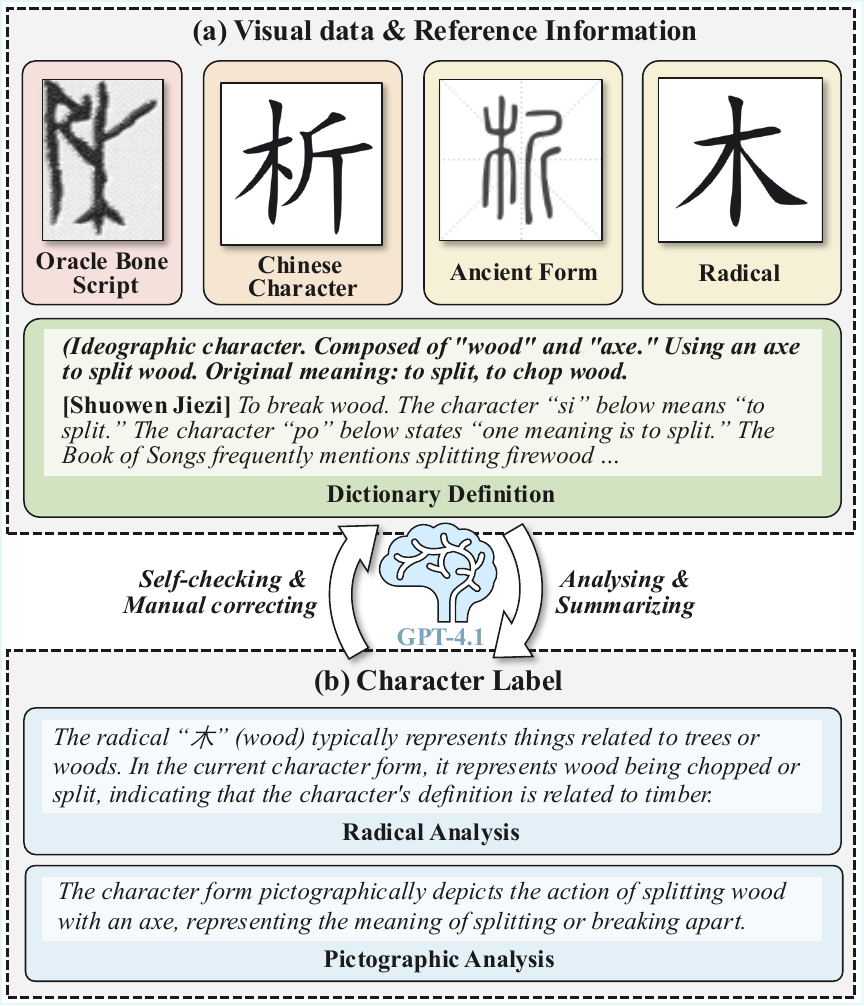}
  \caption{The demonstration of our data engine.}
  \label{fig:dataset}
\end{figure}

As mentioned above, existing LVLMs are still difficult to apply to the OBS deciphering task despite their excellent performance on multiple general tasks. To address this challenge, we introduce the Pictographic Decipherment OBS (PD-OBS) dataset to train LVLMs with the capability for pictographic analysis of OBS, which is of great significance for the OBS decipherment task. The PD-OBS dataset comprises a total of 47,157 Chinese characters. Among these, 3,173 characters are associated with OBS images collected from the public HUST-OBC and EVOBC datasets; 10,968 characters are provided with ancient Clerical Script images from glyph repositories; and all characters are accompanied by modern regular script images from Han Dian. In addition to image data, each character is annotated with radical analysis and pictographic analysis using text, both of which are closely related to the semantic meaning of the character.

The annotation process is conducted in three stages, as illustrated in Figure~\ref{fig:dataset}. First, we retrieve radical tags, definitions, and explanations for each character from Shuowen Jiezi (an ancient Chinese dictionary) via Han Dian. Second, we associate the radical tags obtained and their explanations with modern, ancient clerical, and oracle bone script images of each character. We further utilize GPT-4.1 to enrich the radical tags based on the referenced glyph images and to summarize the analytical content. Finally, both automated self-checking with GPT-4.1 and manual review are performed to correct any labels that are non-standard or deviate from the actual character meanings.

\begin{figure*}[t]
  \centering
  \includegraphics[width=0.96\textwidth]{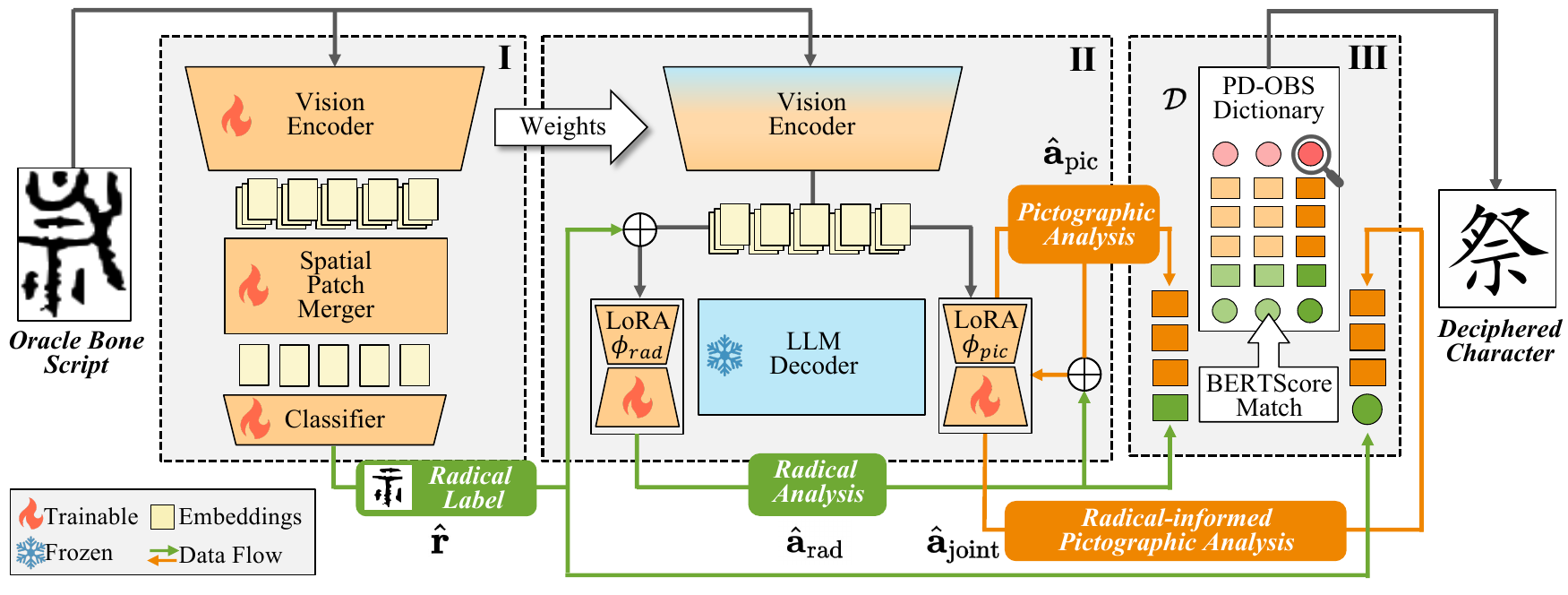}
  \caption{The framework of the proposed method. I is used to indicate the Radical Recognition stage, and II is used to indicate the Radical-Pictographic Mutual Analysis stage, while III is used to indicate Radical-Pictographic Dual Matching.
  }
  \label{fig:framework}
\end{figure*}

This dataset plays a foundational role in the proposed pictographic decipherment framework and is utilized in two key stages of our method:
We construct multi-modal, multi-turn dialogue training samples by pairing OBS images with corresponding modern character labels, enhancing the LVLM's basic capacity to understand OBS glyphs. We group all characters by their radical tags and use a BERT model \cite{bert} to encode the character label text into feature vectors, forming a Chinese character–pictograph analysis dictionary $\mathcal{D}$ that serves as a reference for matching and verifying LVLM-generated decipherment outputs. In summary, the PD-OBS dataset serves as a cornerstone for realizing our LVLM-based pictographic decipherment framework.

\section{Method}
\subsection{Framework}

Our framework is built upon Qwen2.5-VL-7B \cite{qwen2.5}, sharing the same vision encoder and LLM. As illustrated in Figure \ref{fig:framework}, we introduce a spatial patch merger as the visual adapter and a classifier to predict the radical label. We also propose a radical LoRA module $\phi_{\mathrm{rad}}$ and a pictographic LoRA module $\phi_{\mathrm{pic}}$ \cite{lora} to analyze the corresponding information. Furthermore, we design a progressive training—starting from radical recognition, followed by radical and pictographic analysis, and culminating in mutual analysis—to gradually lead the model to OBS decipherment task. In addition, we propose a novel radical-pictographic dual matching mechanism to select the most appropriate character from the database.

\begin{figure}[t]
  \centering
  \includegraphics[width=0.485 \textwidth]{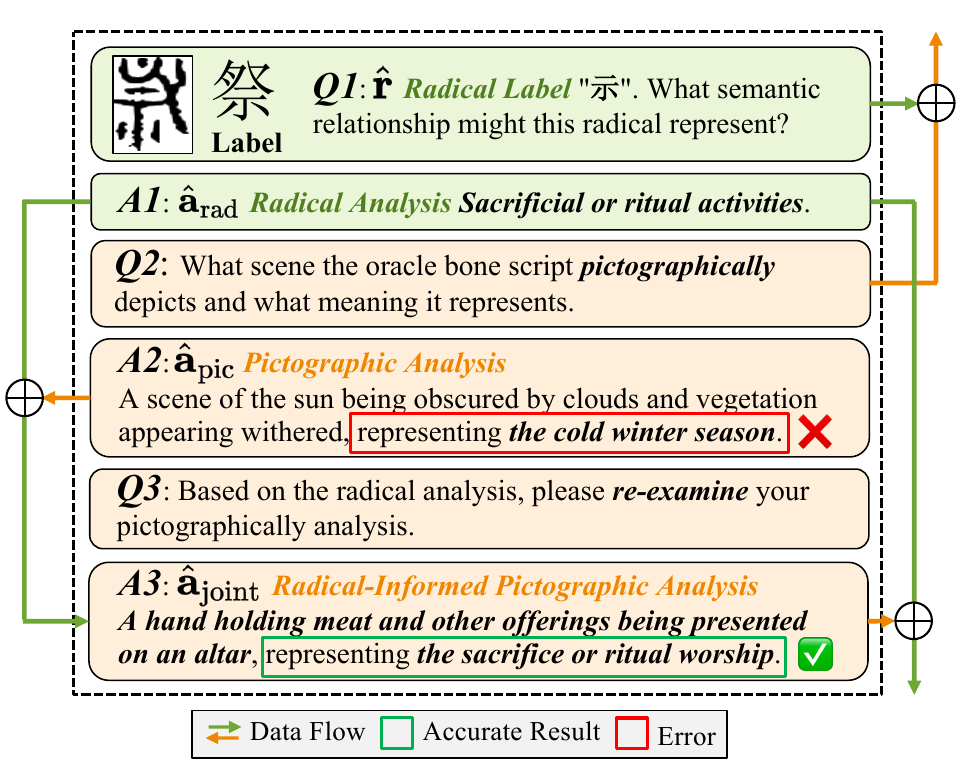}
  \caption{The workflow of proposed radical-pictographic mutual analysis.}
  \label{fig:method}
\end{figure}
\subsection{Radical Recognition} 

In this stage, we aim to adapt the vision encoder to the unique visual style of OBS and to predict radical labels that will serve as critical cues for downstream reasoning. For this purpose, we designed a spatial patch merger as a visual adapter, which compresses and aggregates high-dimensional visual features into a fixed-dimensional feature vector at a predetermined scale, which serves as an abstract representation of the OBS. In addition, we design a triplet loss \cite{tloss} based on the Euclidean distance to explicitly improve the differentiation of feature vectors with different radicals.

Specifically, we implement a sampling strategy to ensure that each batch contains at least two samples for each radical class. During training, for each sample in the batch, we designate its feature vector $V_n$ as an anchor, then select a positive sample $V_n^+$ (\textit{i.e.}, a sample with the same radical label ) and a negative sample  $V_n^-$ (\textit{i.e.}, a sample with a different radical label). The triplet loss is as follows:
\begin{equation}
\mathcal{L}_{\mathrm{trip}} = -\frac{1}{N} \sum_{n=1}^{N} \max\left( \|V_n - V_n^+\|_2 - \|V_n - V_n^-\|_2 + \alpha, \, 0 \right)
\end{equation}
Regarding the classifier, we use the cross-entropy loss $\mathcal{L}_{\mathrm{ce}}$ to optimize it. Therefore, the whole loss function $\mathcal{L}_{\mathrm{stage1}}$  of this stage can be shown as follows:
\begin{equation}
\mathcal{L}_{\mathrm{stage1}}=\gamma \mathcal{L}_{\mathrm{trip}} + \mathcal{L}_{\mathrm{ce}},
\end{equation} where $\gamma$ is a hyperparameter used to balance the two losses.

\subsection{Radical-Pictographic Mutual Analysis} 
To bridge glyphs and meaning in OBS, we design a progressive glyph analysis process to facilitate the decipherment task. Specifically, we introduce a progressive training procedure, beginning with radical analysis, where the radical is predicted in the first stage. In both OBS and ancient Chinese characters, radicals often determine the basic semantic class of a character, as illustrated by the Q1\&A1 in Figure~\ref{fig:method}. Therefore, we train the model’s radical analysis capability with a large number of radical–analysis Q\&A pairs constructed from the PD-OBS dataset. Next, we guide the model to perform a pictographic analysis of the entire character to predict the meaning embedded in the full character glyph, as shown by the Q2\&A2  in Figure~\ref{fig:method}. 

Finally, we design a mutual analysis step to address cases where pictographic analysis alone may not predict the correct corresponding modern character directly. This step informs the pictographic analysis with insights from radical analysis, resulting in more accurate character meanings, as shown by Q3\&A3 in Figure~\ref{fig:method}.

During training, we initialize the model with pretrained visual encoder weights of the previous stage, freezing the shallow layers to retain low-level features while fine-tuning the deeper layers for high-level semantic adaptation. In addition, we introduce a radical LoRA module $\phi_{\mathrm{rad}}$ and a pictographic LoRA module $\phi_{\mathrm{pic}}$ \cite{lora}, the former for radical analysis while the latter for pictographic and mutual analysis. The training data consists of Q\&A pairs from the PD-OBS dataset, as illustrated in Figure~\ref{fig:method}, and the loss function employed is the cross-entropy loss commonly used in LVLM training.

\begin{algorithm}[t]
\small
\caption{Radical-Pictographic Dual Matching}
\label{alg:dual-matching}
\begin{algorithmic}[1]
\REQUIRE Dictionary $\mathcal{D} = \{(\mathbf{r}_i, \mathbf{a}_{\mathrm{rad},i}, \mathbf{a}_{\mathrm{pic},i}, \mathbf{a}_{\mathrm{joint},i}, y_i)\}_{i=1}^N$
\REQUIRE Model output $(\hat{\mathbf{r}}, \hat{\mathbf{a}}_{\mathrm{rad}}, \hat{\mathbf{a}}_{\mathrm{pic}}, \hat{\mathbf{a}}_{\mathrm{joint}})$
\REQUIRE Parameter $k$ (top-$k$)
\REQUIRE $\mathcal{S}(\cdot, \cdot)$: semantic similarity between text sequences computed by BERT-Score~\cite{bert}

\STATE // Filtered Matching
\STATE $\mathcal{D}_{\mathrm{rad}} \leftarrow \{i \mid \mathbf{r}_i = \hat{\mathbf{r}}\}$
\STATE $C_1 \leftarrow \text{top-}k$ indices in $\mathcal{D}_{\mathrm{rad}}$ by $\mathcal{S}(\mathbf{a}_{\mathrm{pic},i}, \hat{\mathbf{a}}_{\mathrm{pic}})$

\STATE // Joint Matching
\STATE $C_2 \leftarrow \text{top-}k$ indices in $\{1,\ldots,N\}$ by $\mathcal{S}((\mathbf{a}_{\mathrm{rad},i} \oplus \mathbf{a}_{\mathrm{joint},i}), (\hat{\mathbf{a}}_{\mathrm{rad}} \oplus \hat{\mathbf{a}}_{\mathrm{joint}}))$

\STATE // Merge and Rerank
\STATE $C \leftarrow C_1 \cup C_2$
\STATE $R \leftarrow \text{top-}k$ in $C$ by their similarity scores

\RETURN $\{y_i \mid i \in R\}$
\end{algorithmic}

\end{algorithm}

\subsection{Radical-Pictographic Dual Matching} 
Following the first two stages, we obtain four intermediate results for each test character: predicted radical label $\hat{\mathbf{r}}$, radical analysis $\hat{\mathbf{a}}_{\mathrm{rad}}$, pictographic analysis $\hat{\mathbf{a}}_{\mathrm{pic}}$, and radical-informed pictographic analysis $\hat{\mathbf{a}}_{\mathrm{joint}}$. We propose a dictionary-based dual matching mechanism for decipherment. Given the candidate dictionary $\mathcal{D} = \{(\mathbf{r}_i, \mathbf{a}_{\mathrm{rad},i}, \mathbf{a}_{\mathrm{pic},i}, \mathbf{a}_{\mathrm{joint},i}, y_i)\}_{i=1}^N$ from the PD-OBS dataset, the mechanism works as follows:

First, we filter candidates by the predicted radical label $\hat{\mathbf{r}}$, then select the top-$k$ entries by the semantic similarity $\mathcal{S}(\mathbf{a}_{\mathrm{pic},i}, \hat{\mathbf{a}}_{\mathrm{pic}})$ (BERT-Score) \cite{bert} between the pictographic analysis.  
Second, we concatenate the radical analysis and radical-informed pictographic analysis, and select another top-$k$ entries by similarity $\mathcal{S}((\mathbf{a}_{\mathrm{rad},i} \oplus \mathbf{a}_{\mathrm{joint},i}), (\hat{\mathbf{a}}_{\mathrm{rad}} \oplus \hat{\mathbf{a}}_{\mathrm{joint}}))$.  
Finally, we merge and rerank these candidate sets to obtain the top-$k$ modern Chinese characters as decipherment results.  
All steps and notations are detailed in Algorithm~\ref{alg:dual-matching}.

Notably, we employ the matching mechanism instead of directly outputting decipherment results, which helps mitigate the limited generalization of the model for zero-shot settings and undeciphered OBS caused by the absence of such OBS in the training data.

\section{Experiments}
\subsection{Implementation Details}
All training and evaluation experiments are conducted on 8 NVIDIA RTX 4090 GPUs. We initialize our model with the pretrained weights of Qwen2.5-VL-7B. During the radical recognition stage, we set the learning rate to 5e-4, batch size to 8, and train for 5 epochs. For the pictographic decipherment stage, we use a learning rate of 5e-5, batch size of 4, and train for 4,000 steps. AdamW \cite{adamw} is used as the optimizer. The radical LoRA $\phi_{\mathrm{rad}}$ and pictographic LoRA $\phi_{\mathrm{pic}}$ are configured with a dropout rate of 0.05 and 0.25 respectively, and both use rank and alpha values of 32.

\begin{table}[t]
\centering
\renewcommand{\arraystretch}{1.15}
\setlength{\tabcolsep}{4pt}
\small

\begin{tabular}{
    >{\raggedright\arraybackslash}p{1.45cm} 
    >{\centering\arraybackslash}p{1.4cm} 
    >{\centering\arraybackslash}p{1.32cm} 
    >{\centering\arraybackslash}p{1.4cm} 
    >{\centering\arraybackslash}p{1.5cm}
}
\toprule
\multirow{3}{*}{\textbf{Method}} &
\multicolumn{2}{c}{\textbf{Validation}} &
\multicolumn{2}{c}{\textbf{Zero-shot}} \\
& \makecell[c]{HUST-\\OBC} & \makecell[c]{EV-OBC} & \makecell[c]{HUST-\\OBC} & \makecell[c]{EV-OBC} \\
\midrule
\multicolumn{5}{l}{\textit{classification model-based}} \\
InceptionV3 & 74.4, 76.9 & 62.4, 64.5 & - & - \\
ViT & 79.2, 81.7 & 72.7, 74.2 & - & - \\
PyGT & \textbf{84.3}, \underline{87.6} & \textbf{78.1}, \underline{81.2} & - & - \\
\midrule
\multicolumn{5}{l}{\textit{Commercial LVLM}} \\
GPT-4.1 & 6.0, 6.0 & 4.5, 4.5 & 5.3, 5.3 & 4.3, 4.3 \\
QwenVLMax & 4.8, 4.8 & 4.1, 4.1 & 2.0, 2.0 & 4.0, 4.0 \\
\midrule
\multicolumn{5}{l}{\textit{Diffusion-based}} \\
OBSD & 66.8, 72.9 & 71.2, 77.9 & \textbf{18.3}, \underline{27.5} & \underline{30.4}, \underline{50.5} \\
BBDM & 55.8, 59.5 & 60.3, 62.1 & 8.0, 14.1 & 19.5, 29.5 \\
\midrule
\textbf{Ours} & \underline{80.6}, \underline{87.8} & \underline{76.3}, \textbf{81.7} & \underline{16.8}, \textbf{53.7} & \textbf{33.3}, \textbf{64.1} \\

\textit{Improvement} & -3.7, +0.2 &  -1.8, +0.5 & -1.5, +26.2 & +2.9, +13.6\\

\bottomrule
\end{tabular}
\small
\caption{Each cell reports Top-1 (left) and Top-10 (right) accuracy (in \%). The best and second-best results are respectively marked in bold and underlined. \textit{Improvement} represents the improvement achieved by our method compared to the existing best method.}
\label{tab:obc-comparison}

\end{table}
\begin{table}[ht]

\centering

\renewcommand{\arraystretch}{1.15}
\setlength{\tabcolsep}{8pt}
\begin{tabular}{
    >{\raggedright\arraybackslash}p{2.56cm} 
    >{\centering\arraybackslash}p{0.8cm} 
    >{\centering\arraybackslash}p{0.8cm} 
    >{\centering\arraybackslash}p{0.8cm} 
    >{\centering\arraybackslash}p{0.8cm}
}
\toprule
\multirow{2}{*}{\textbf{Method}}& \multicolumn{2}{c}{\textbf{HUST-OBC}} & \multicolumn{2}{c}{\textbf{EVOBC}} \\
\cmidrule(lr){2-3} \cmidrule(lr){4-5}
& \textit{Valid.} & \textit{ZS} & \textit{Valid.} & \textit{ZS} \\
\midrule
Qwen2.5-VL-7B  & 69.4 & 65.1 & 68.3 & 67.9 \\
Qwen-VL-Max    & 70.5 & 65.6 & 69.8 & 68.2 \\
GPT-4.1        & \underline{73.7} & \underline{67.5} & \underline{71.4} & \underline{70.9} \\

\midrule
Ours           & \textbf{94.6} & \textbf{79.4} & \textbf{93.7} & \textbf{84.9} \\
\bottomrule
\end{tabular}
\caption{The Valid. and ZS indicate validation and zero-shot settings, respectively. The Bert-Score (in \%) achieved by different methods.}
\label{tab:explain}

\end{table}

\subsection{DataSets and Evaluation Metrics}
We perform quantitative evaluations on the HUST-OBC \cite{hust-obs} and EV-OBC \cite{evobc} datasets, selecting 200 character classes from each as zero-shot test sets. The remaining data are randomly split into training and validation sets in a 9:1 ratio to assess the OBS recognition capabilities of our framework and baselines. We use Top-k accuracy as evaluation metrics as in previous work \cite{obsd,PYGT,oraclesage,obi-bench}, which is usually used in diverse classification tasks \cite{ViT,MAAN_cnn-based,orerf-cnn_based,2dl_cnn,swin_cnn-based,survey, survey2, survey3}.

\subsection{Main Results} 
To evaluate the effectiveness of our approach on OBS decipherment, we conduct comprehensive comparisons on two benchmark datasets, HUST-OBC \cite{hust-obs} and EV-OBC \cite{evobc}, under both validation and zero-shot settings, as shown in Table \ref{tab:obc-comparison}. We adopt InceptionV3 \cite{inception-v3}, ViT \cite{ViT}, and PyGT \cite{PYGT} as classification model-based baselines, and OBSD \cite{obsd} and BBDM \cite{bbdm} as diffusion-based methods. Due to the lack of open-source implementations and dataset inconsistency, existing composition-based methods \cite{ppp_composition-based,retrieval_composition-based,seg_composition-based}  are not considered for inclusion in the comparison methods for the time being. Instead, we include strong commercial LVLMs, GPT-4.1 and QwenVLMax, for comparison. In contrast, commercial LVLMs perform poorly in both settings, with Top-1 accuracy consistently below 6\%, highlighting their limited capability to understand the visual structure of ancient scripts. On the validation set, although our method yields slightly lower Top-1 accuracy than the best classification model-based baseline (e.g., PyGT), it achieves the highest Top-10 accuracy, demonstrating superior capability in generating high-quality candidates and offering greater practical utility. In the more challenging zero-shot scenario, our method exhibits notably strong performance: It remains competitive in Top-1 accuracy and significantly outperforms all methods in Top-10 accuracy, surpassing the second-best method by 26.2\% on HUST-OBC and 13.6\% on EV-OBC. These results confirm the strong generalization and transferability of our method to unseen OBS, highlighting its potential value in assisting the recognition of undeciphered OBS in archaeological research.

To quantitatively evaluate the precision of radical and pictographic analysis produced by our method, we employ BERT-Score \cite{bert} to measure the similarity between the top-1 outputs and the ground-truth labels from the dictionary $\mathcal{D}$. We also evaluate other LVLMs, including GPT-4.1, Qwen-VL-Max, and Qwen2.5-VL-7B, and compare their average BERT-Score on both the validation and zero-shot settings of the HUST-OBC \cite{hust-obs} and EVOBC \cite{evobc} datasets. As shown in Table~\ref{tab:explain}, our method significantly outperforms the state-of-the-art LVLM GPT-4.1 21.60\% and 12.95\%, averaged over the two datasets, under validation and zero-shot settings, respectively. This result indicates that the analysis generated by our framework are more reliable and informative.

\begin{figure*}[t]
  \centering
  \includegraphics[width=0.97175\textwidth]{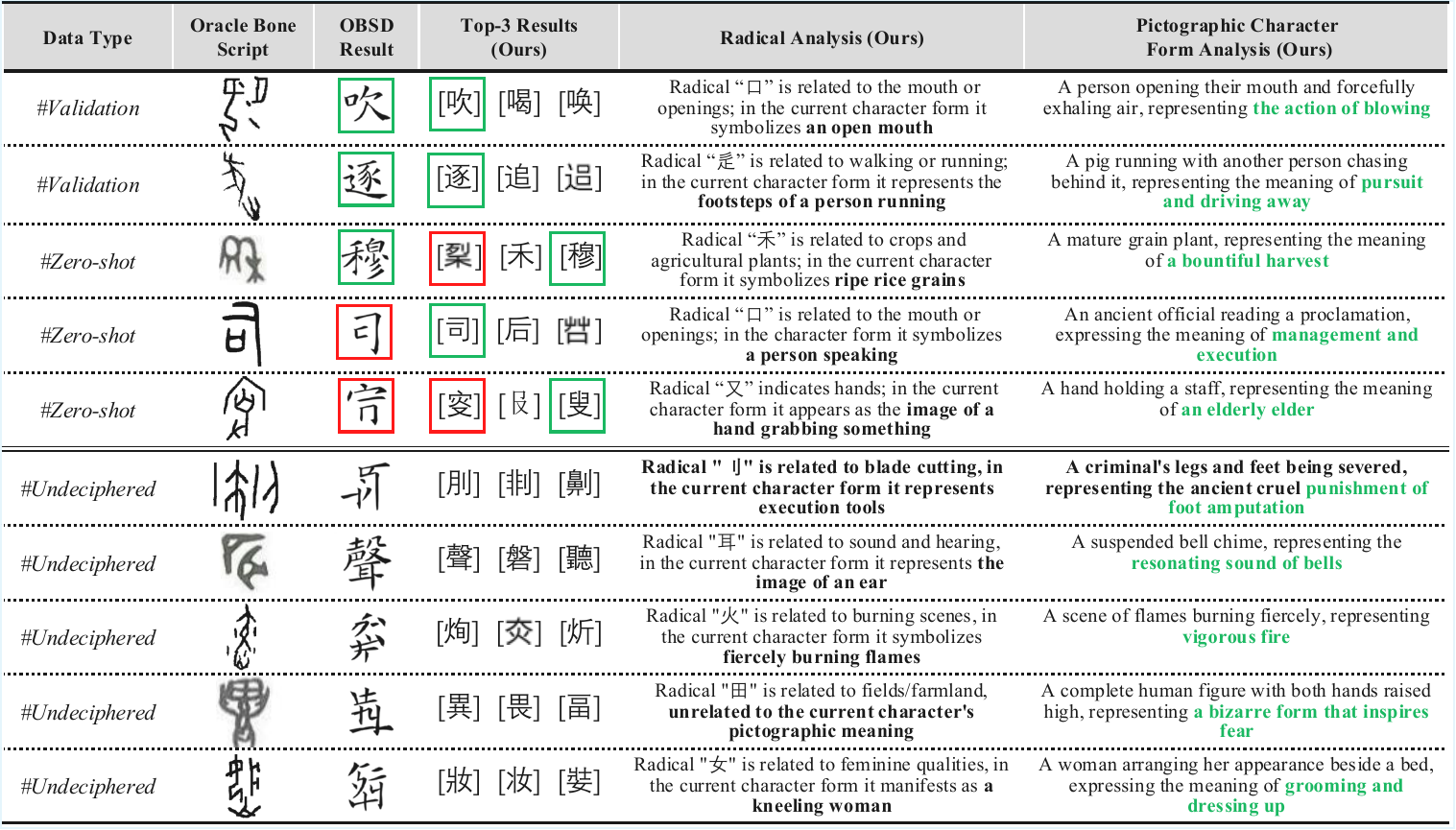}
  \caption{Visualization of the deciphering process and results, under the settings of validation, zero-shot, and undeciphered characters.
  }
  \label{fig:vis-2}
\end{figure*}

\subsection{Ablation Study}
\begin{table}[t]
\centering
\renewcommand{\arraystretch}{1.15}
\setlength{\tabcolsep}{6pt}

\begin{tabular}{
    >{\raggedright\arraybackslash}m{3.88cm}
    >{\centering\arraybackslash}m{1.3cm}
    >{\centering\arraybackslash}m{1.8cm}
}
\toprule
\textbf{Method} & \textbf{Validation} & \textbf{Zero-shot} \\
\midrule
Vision Encoder of Qwen & 92.7 & 87.1 \\
+ Our Radical Recognition
 & 93.6 & 88.3 \\
+ LoRA-based Recognition & 80.1 & 69.8 \\
\bottomrule
\end{tabular}
\caption{The performance of radical recognition accuracy (in \%) of the variants of our model.}
\label{tab:vision-zeroshot}

\end{table}

To evaluate the effectiveness of the proposed radical recognition stage, we use the original vision encoder of Qwen2.5-VL-7B \cite{qwen2.5} as the baseline, and incorporate our radical recognition module or a LoRA-based recognition method. Their recognition accuracy is reported on the HUST-OBS dataset \cite{hust-obs} with validation and zero-shot settings. Our method introduces a spatial patch merger and the loss function $\mathcal{L}_{\mathrm{trip}}$  on top of the baseline vision encoder, resulting in improvements of 0.9\% and 1.2\% accuracy on the validation and zero-shot settings, respectively. The LoRA-based recognition method merges the recognition stage with the radical analysis process and training with LoRA-based fine-tuning. The results demonstrate that this method leads to a significant drop in radical recognition accuracy and introduces substantial errors in radical analysis, thus we retain radical recognition as an independent stage in our framework.
\begin{table}[t]
\centering
\small
\renewcommand{\arraystretch}{1.15}
\setlength{\tabcolsep}{2pt}

\begin{tabular}{
    >{\raggedright\arraybackslash}p{4.23cm} 
    >{\centering\arraybackslash}p{0.91cm} 
    >{\centering\arraybackslash}p{0.94cm} 
    >{\centering\arraybackslash}p{0.91cm} 
    >{\centering\arraybackslash}p{0.94cm}
}
\toprule
\multirow{2}{*}{\textbf{Method}} & \multicolumn{2}{c}{\textbf{Validation}} & \multicolumn{2}{c}{\textbf{Zero-shot}} \\
\cmidrule(lr){2-3} \cmidrule(lr){4-5}
 & Top-1 & Top-10 & Top-1 & Top-10 \\
\midrule
\makecell[l]{QWen2.5-VL-7B} & 1.4 & 1.4 & 0.2 & 0.2 \\
\makecell[l]{+ LoRA} & 52.4 & 52.4 & 1.6 & 1.6 \\
 \makecell[l]{+ Rad\&Pic Mutual Analysis} & 60.3 & 61.4 & 5.2 & 5.4 \\
 \makecell[l]{+ Radical Recognition} & 64.2 & 64.2 & 6.6 & 6.6 \\
 \makecell[l]{+ Rad\&Pic Dual Matching(Ours)} & 80.6 & 87.8 & 16.8 & 53.7 \\
\bottomrule
\end{tabular}
\caption{Top-1 and Top-10 accuracy ( in \%) of our model and its variants on HUST-OBS dataset.}
\label{tab:qwen25-ablation}

\end{table}

To validate the effectiveness of our proposed modules and strategies, we take Qwen2.5-VL-7B \cite{qwen2.5} as the baseline and incrementally add each component to form our final model. The Top-1 and Top-10 performance under both validation and zero-shot settings are shown in Table \ref{tab:qwen25-ablation}. The results demonstrate that LoRA fine-tuning (+LoRA) enables initial decipherment ability on the validation set but still lacks generalization in zero-shot scenarios. With the introduction of Radical-Pictographic Mutual Analysis and Radical Recognition, the model’s accuracy continues to improve on the validation set, but the increase in zero-shot ability is still very limited. The primary cause lies in the insufficient generalization capability of the model trained via LoRA-based supervised fine-tuning, which often fails to generate rare characters—a common challenge in zero-shot scenarios. To mitigate this, we introduce the Radical-Pictographic Dual Matching mechanism as a replacement for direct prediction. This strategy not only significantly improves the model’s zero-shot performance but also enhances the robustness of the OBS whose radicals are not related to semantics, ensuring reliable and verifiable decipherment results.

\subsection{Qualitative Results}

Figure~\ref{fig:vis-2} presents qualitative results of our method and OBSD \cite{obsd} under three settings: validation, zero-shot, and undeciphered OBS. As shown, our model demonstrates strong recognition capabilities on the validation set and also generalizes well to unseen OBS in zero-shot settings. More notably, for characters that remain undeciphered by human experts, the model is able to produce semantically plausible predictions, accompanied by interpretable analysis. Our design of Radical-Pictographic Mutual Analysis plays a key role: The radical analysis component traces the structural origins of radicals and explains their symbolic function in the current character form. Meanwhile, the pictographic form analysis provides a holistic visual-semantic mapping based on the character's overall shape and implied meaning. Together, these complementary analysis form a dual reasoning path that enhances the model's ability to generate semantically grounded and interpretable outputs, even for previously untranscribed scripts.

\section{Conclusion}

We propose an interpretable OBS decipherment framework through radical and pictographic analysis. The framework bridges glyph to meaning through three stages: radical recognition and analysis, pictographic analysis, and mutual analysis. With the proposed radical-pictographic dual matching, our model can filter the appropriate deciphering candidate set from the dictionary based on analysis results, replacing direct output of deciphering results to achieve better zero-shot performance. Moreover, the generated textual analysis serve as interpretable content, offering references for undeciphered OBS characters, thus holding great potential for archaeological applications. To support training, we construct the PD-OBS dataset, comprising 47,157 Chinese characters annotated with OBS images and pictographic analysis texts, providing a valuable resource for future research. Experimental results demonstrate that our method achieves strong performance in decipherment accuracy, generalization, and interpretability.

\newpage
\bibliography{aaai2026}

\clearpage

\clearpage
\twocolumn[
\begin{center}
    \LARGE\bf Supplementary Materials
\end{center}
\vspace{1.2cm}
]

\section{Limitations and Future Work}

The primary limitations of our proposed decipherment framework are manifested in insufficient generalization capability and degraded fundamental reasoning ability resulting from the LoRA training approach. We observe that the oracle bone script dataset contains numerous characters with similar glyphs or semantics, leading the model to rely on similar character information from training labels in zero-shot testing scenarios rather than conducting thorough radical and pictographic analysis, consequently causing deviations in results. For example, when the training set involves the Chinese character "pin" composed of three "kou" radicals, and the zero-shot test contains the Chinese character "ji" composed of four "kou" radicals, the model sometimes skips the pictographic analysis process and directly outputs the label of the Chinese character "pin", because these two characters have highly similar glyphs and meanings.

To address these limitations, future improvements will consider applying state-of-the-art reinforcement learning frameworks and targeted reward function to further overcome the model's generalization constraints. Additionally, we will attempt to integrate composition-based methods to enhance the model's robustness for semantically complex yet structurally well-defined characters.

\section{Parameter Analysis in RDM}

\subsection{Top-k Parameter Analysis}

As shown in Table \ref{tab:top-k}, with the increase of Top-k, the zero-shot accuracy of our method rises significantly, achieving considerable accuracy within Top-10. This result demonstrates that Radical-pictographic Dual Matching (RDM) mechanism can accurately locate the deciphered character sets, overcome the confusion problem of Chinese characters with similar structures or meanings, and effectively improve the generalization. A small-scale increase in Top-k has minimal impact on the difficulty of textual research, while bringing significant accuracy improvements, which demonstrates the application value of our decipherment framework in practical archaeological decipherment scenarios, providing concise and reliable decipherment analysis references.

\subsection{Dictionary Scale Study}

In practice, only a limited subset of over 100,000 Chinese characters has achieved widespread circulation and possesses well-defined meanings, thus, selecting an appropriate candidate set for oracle bone script decipherment is crucial. Under the PD-OBS dataset setting, the matching dictionary comprises 47,157 Chinese characters documented in the Kangxi Dictionary (a Chinese dictionary). 

\begin{table}[ht]
\centering
\small
\renewcommand{\arraystretch}{1.15}
\setlength{\tabcolsep}{8pt}
\resizebox{0.48\textwidth}{!}{%
\begin{tabular}{
    >{\raggedright\arraybackslash}p{2.56cm} 
    >{\centering\arraybackslash}p{0.8cm} 
    >{\centering\arraybackslash}p{0.8cm} 
    >{\centering\arraybackslash}p{0.8cm} 
    >{\centering\arraybackslash}p{0.8cm}
}
\toprule
\multirow{2}{*}{\textbf{@Top-k}}& \multicolumn{2}{c}{\textbf{HUST-OBC}} & \multicolumn{2}{c}{\textbf{EVOBC}} \\
\cmidrule(lr){2-3} \cmidrule(lr){4-5}
& \textit{Valid.} & \textit{ZS} & \textit{Valid.} & \textit{ZS} \\
\midrule
Top-1    & 80.6 & 16.8 & 76.3 & 33.3 \\
Top-5    & 86.0 & 39.3 & 79.8 & 56.0 \\
Top-10   & 87.8 & 53.7 & 81.7 & 64.1 \\
Top-50   & 92.1 & 74.2 & 88.0 & 80.2 \\
Top-100  & 94.4 & 82.5 & 91.2 & 89.7 \\
\bottomrule
\end{tabular}%
}
\caption{Decipherment accuracy (in \%) under different Top-k settings.The Valid. and ZS indicate validation and zero-shot settings, respectively.}
\label{tab:top-k}
\end{table}

\begin{table}[ht]
\centering
\small
\renewcommand{\arraystretch}{1.15}
\setlength{\tabcolsep}{8pt}
\resizebox{0.48\textwidth}{!}{%
\begin{tabular}{
    >{\raggedright\arraybackslash}p{2.56cm} 
    >{\centering\arraybackslash}p{0.8cm} 
    >{\centering\arraybackslash}p{0.8cm} 
    >{\centering\arraybackslash}p{0.8cm} 
    >{\centering\arraybackslash}p{0.8cm}
}
\toprule
\multirow{2}{*}{\textbf{@Dict. Scale}}& \multicolumn{2}{c}{\textbf{HUST-OBC}} & \multicolumn{2}{c}{\textbf{EVOBC}} \\
\cmidrule(lr){2-3} \cmidrule(lr){4-5}
& \textit{Valid.} & \textit{ZS} & \textit{Valid.} & \textit{ZS} \\
\midrule
7000    & 59.6 & 31.9 & 65.7 & 48.2 \\
10000   & 73.7 & 39.3 & 79.5 & 59.8 \\
20902   & 86.5 & 51.8 & \textbf{83.9} & \textbf{66.4} \\
27928   & \textbf{88.3} & \textbf{54.1} & 82.2 & 64.5 \\
47157   & 87.8 & 53.7 & 81.7 & 64.1 \\
\bottomrule
\end{tabular}%
}
\caption{Decipherment top-10 accuracy (in \%) under different dictionary scales.The Valid. and ZS indicate validation and zero-shot settings, respectively.}
\label{tab:dict}
\end{table}
We constructed four additional subset dictionaries: 7,000 commonly used Chinese characters, 10,000 commonly used characters including known oracle bone script decipherment results, 20,902 Unicode-supported characters, and 27,928 characters encompassing known oracle bone script decipherment results and Unicode Extension A characters. We evaluated the top-10 accuracy of our decipherment framework under different dictionary scales.

As demonstrated in Table \ref{tab:dict}, larger candidate dictionaries do not necessarily yield superior results; instead, candidate dictionaries with scales ranging from 20,000 to 30,000 characters prove more suitable. This indicates a trade-off relationship between decipherment accuracy and potential recall rate. To ensure the reliability and authority of decipherment results, we employ all characters documented in the Kangxi Dictionary as the PD-OBS dictionary for comparative experiments in both Main Results and Ablation Studies.
\begin{figure*}[t]
  \centering
  \includegraphics[width=0.971\textwidth]{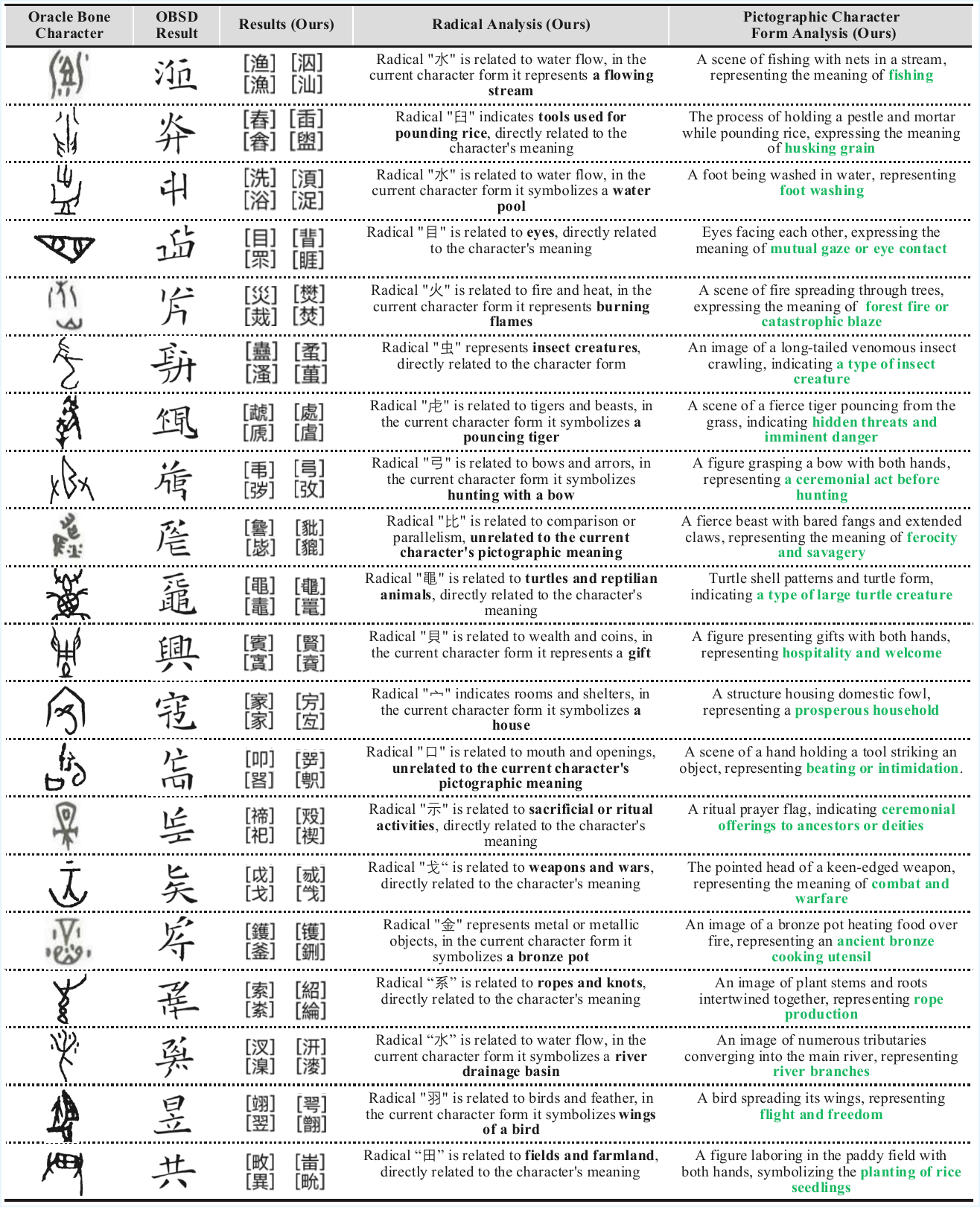}
  \caption{Visualization of the results and deciphering processes of undeciphered characters.
  }
  \label{fig:ud}
\end{figure*}
\section{ Extended Qualitative Results}

\begin{figure*}[t]
  \centering
  \includegraphics[width=0.971\textwidth]{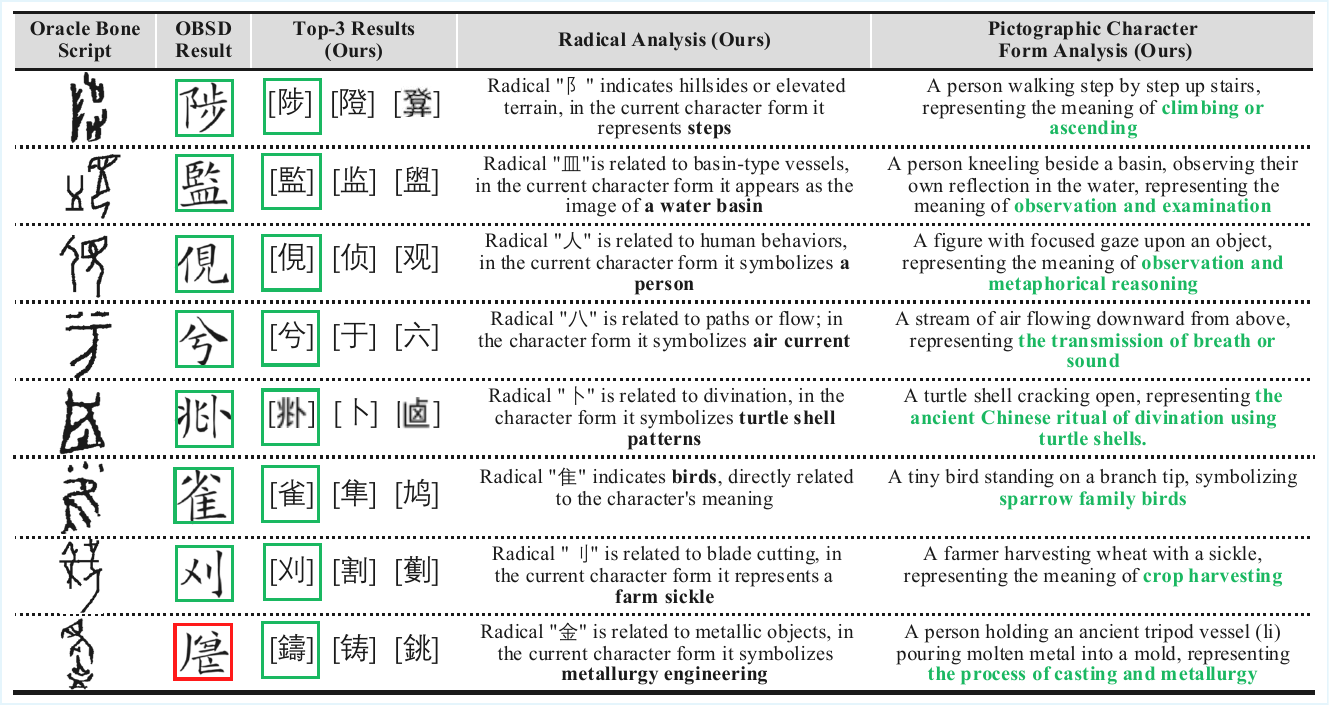}
  \caption{Visualization of the results and deciphering processes of characters in validation setting.
  }
  \label{fig:val}
\end{figure*}

\begin{figure*}[t]
  \centering
  \includegraphics[width=0.971\textwidth]{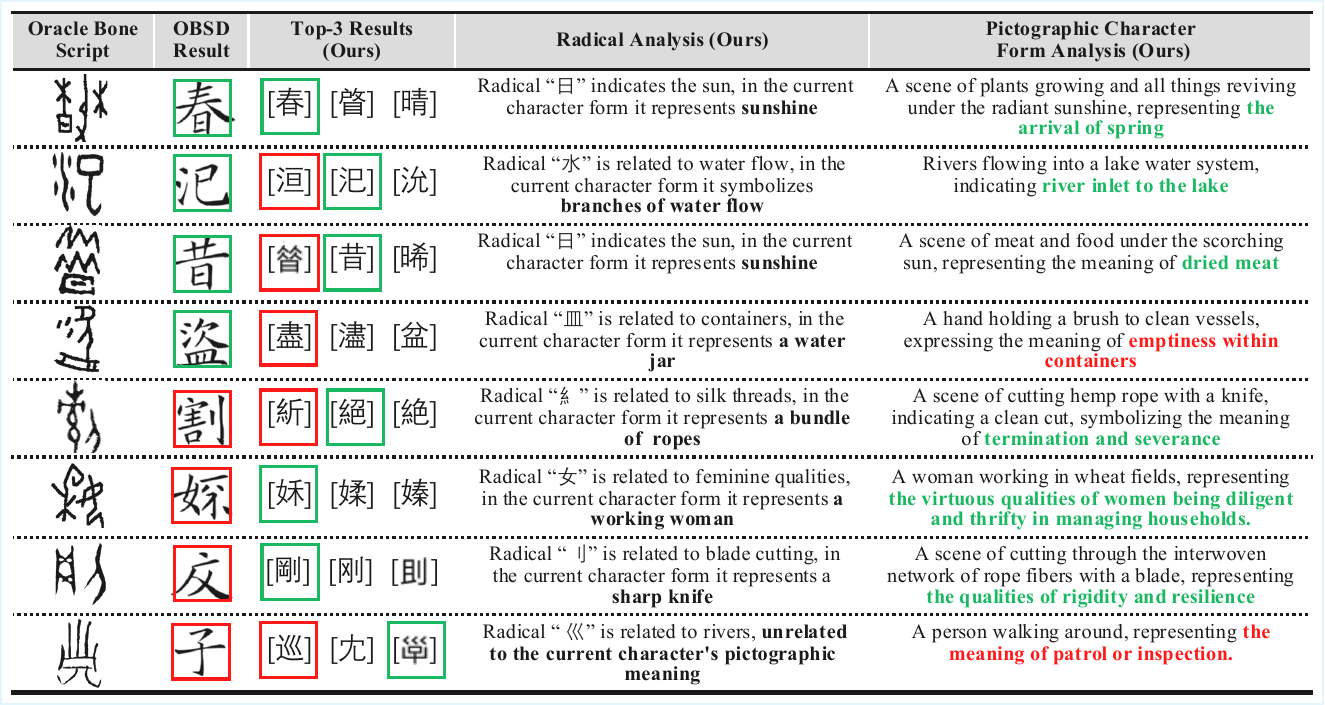}
  \caption{Visualization of the results and deciphering processes of characters in zero-shot setting.
  }
  \label{fig:zs}
\end{figure*}

Figure \ref{fig:val}, Figure \ref{fig:zs} and Figure \ref{fig:ud} visualize additional results and deciphering processes from our proposed framework and OBSD. Moving forward, we will publicly release decipherment results for all undeciphered oracle bone characters, along with the PD-OBS dataset, model weights, and complete codes. 

\end{document}